\documentclass[10pt,twocolumn,letterpaper]{article}

\usepackage[pagenumbers]{wacv} 

\definecolor{wacvblue}{rgb}{0.21,0.49,0.74}
\usepackage[pagebackref,breaklinks,colorlinks,allcolors=wacvblue]{hyperref}

\def\wacvPaperID{1248} 
\def\confName{WACV}
\def\confYear{2026}

\usepackage{makecell}
\usepackage{multirow}
\usepackage{multicol}
\usepackage[table]{xcolor}
\newcommand{\myheading}[1]{\noindent{\textbf{#1}}}

\usepackage{soul}

\title{Prepare Lighter and Faster for Open-Vocabulary Queries: A Query-Wise 3D Segmenter for Gaussian Splatting}

\author{
Nhat-Quynh Le-Pham\footnotemark[1]\thanks{Work done while at Qualcomm.}\\
{\tt\small qule@tcd.ie}
\and
Khoi Nguyen\\
{\tt\small khoi@qti.qualcomm.com}
\and
Binh-Son Hua\\
{\tt\small binhson.hua@tcd.ie}
\and
Qualcomm AI Research\thanks{Qualcomm AI Research is an initiative of Qualcomm Technologies, Inc.}
}

\begin{document}

\maketitle

\begin{abstract}
Open-vocabulary 3D understanding aims to localize and segment arbitrary objects beyond predefined categories. However, existing approaches based on Gaussian Splatting often rely on expensive preprocessing and lengthy training, limiting their applicability when fast and lightweight solutions are desired. We propose a \emph{lightweight, training-free framework} for query-driven 3D instance segmentation on Gaussian Splatting. Instead of conventional mask-wise CLIP embeddings, we use dense pixel-level CLIP features to preserve fine-grained semantic information and enable flexible text-driven localization without relying on predefined instance proposals. Given a text query, we identify a high-quality target region in a carefully selected pivot view and propagate it across views using a video segmentor, providing consistent and comprehensive coverage of the target instance. We then lift the resulting 2D tracklets to 3D Gaussians through a novel visibility voting strategy, yielding accurate 3D segmentation masks with clean object boundaries. Extensive experiments demonstrate state-of-the-art performance on LERF-OVS, ScanNet, and ScanNet++, competitive results on 3D-OVS, and significant speedups over existing open-vocabulary 3D segmentation methods. These results highlight the effectiveness of our framework for efficient open-vocabulary 3D understanding without per-scene training.
\end{abstract}
  
\section{Introduction}
\label{sec:intro}
\begin{figure*}[t]
    \centering
    \includegraphics[width=0.85\linewidth]{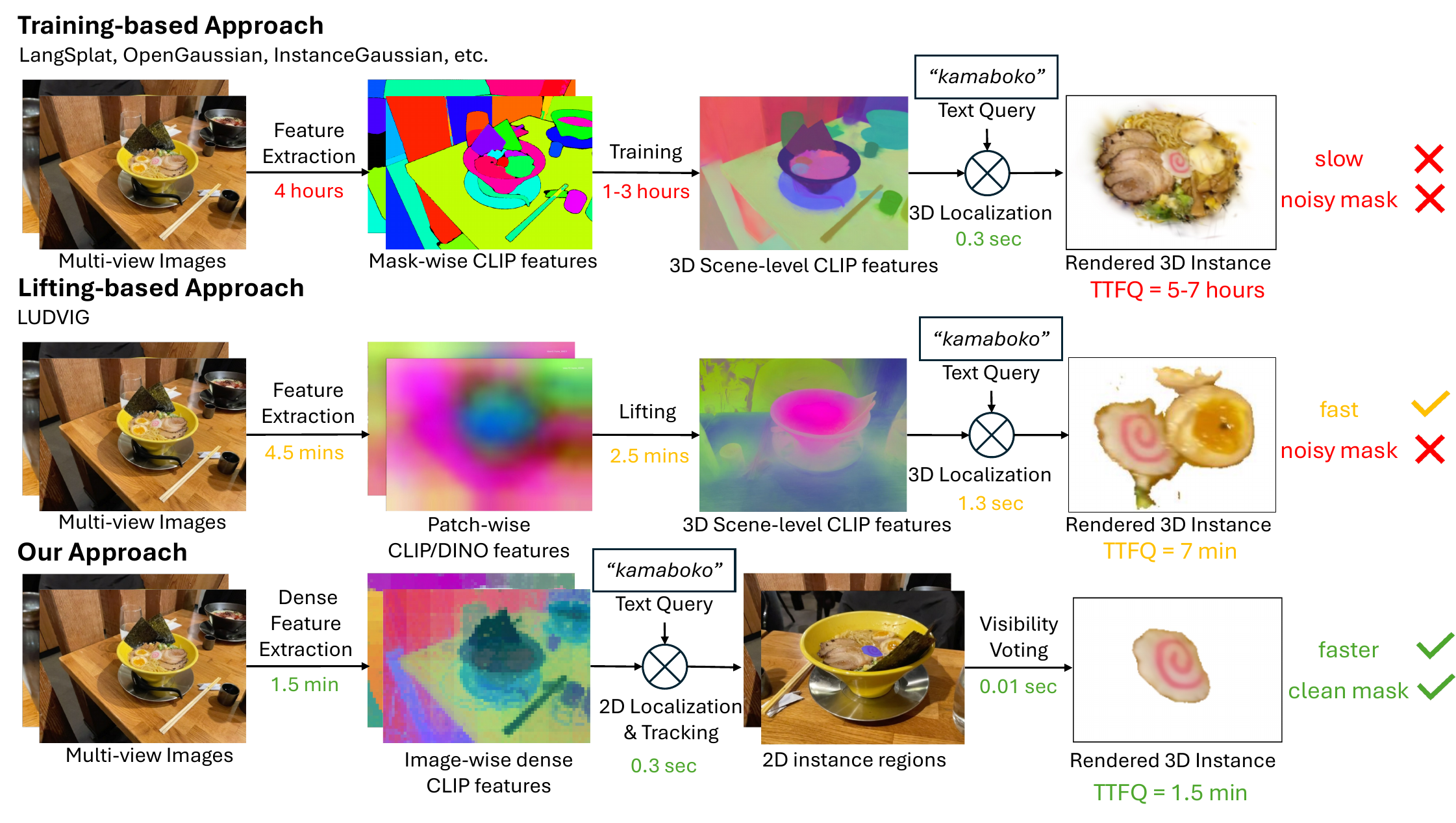}
    \vspace{-10pt}
    \caption
    {\textbf{Time-to-First-Query (TTFQ) comparison on the ``ramen'' scene from LERF-OVS.} Given a 3DGS scene, our method enables 3D instance segmentation after only 1.5 minutes of feature extraction, while preserving low query latency and supporting real-time rendering of clean and accurate segmentation masks directly from Gaussian rasterization. Our method differs from previous methods in three key aspects: processing time, training requirement, and mask quality. Compared to both training-based and lifting-based approaches, our method eliminates the training requirement for integrating semantic features to 3D Gaussians. We instead adopt pixel-level CLIP features to create pivot masks for fast object localization, and then use a tracking segmentor to propagate masks for lifting to 3D via a lightweight voting mechanism, enabling accurate semantic rasterization from 3DGS while significantly optimizing TTFQ time.}
    \label{fig:teaser}
    \vspace{-10pt}
\end{figure*}
Open-vocabulary 3D understanding enables recognition and segmentation of arbitrary objects beyond fixed taxonomies, supporting applications such as robotic perception, AR/VR, and embodied AI.
Gaussian Splatting methods \cite{2dgs, 3dgs} have emerged as efficient approaches for representing high-quality 3D scenes while enabling real-time rendering. Recent advances in fast optimization and feed-forward 3D Gaussian Splatting \cite{yonosplat, anysplat, instantsplat, noposplat, fastgs} enable the reconstruction of 3D scenes from multi-view images within seconds to minutes.
The widespread adoption of 3D Gaussian representations has made sharing pretrained 3D Gaussian assets across teams a common practice. However, despite this progress, efficient open-vocabulary interaction via \textbf{open-vocabulary 3D segmentation} with pretrained 3D Gaussian scenes remains underexplored. 

Most existing solutions \cite{langsplat, legaussian, opengaussian, instancegaussian, langsplatv2, liang_supergseg_2024, saga, panogs, opensplat3d} adopt \textit{training-based} pipelines that integrate vision-language features from VFMs into Gaussian-based 3D scene representations as an additional training stage after the 3D Gaussian primitives are learned (see \cref{fig:teaser}). These methods distill language/instance features into Gaussian splats through multi-stage contrastive learning or feature fusion. In practice, they extract thousands of multi-granularity SAM masks and corresponding mask-level CLIP embeddings across all training views, followed by feature distillation or fusion over these masks and post-processing steps to refine instance-level representations. Although these approaches demonstrate open-vocabulary scene understanding, their preprocessing and training stages are computationally expensive and can take hours to process a scene, making them impractical for instant usage.

Recent efforts \cite{ludvig} have explored \textit{lifting-based} alternatives that partially alleviate preprocessing and training bottlenecks by directly lifting vision foundation model (VFM) features into 3D Gaussian splats. While this approach significantly accelerates scene-level understanding, the lifting process itself still requires over 10 minutes per scene to fuse features across the entire scene, even when only specific objects or regions are of interest. This leads to redundant computation and memory overhead. Moreover, this method exhibits a relatively high query latency of over 1.3 seconds per query, and the resulting 3D masks often lack precise object boundaries, still requiring additional SAM-based refinement. Both paradigms redundantly encode full-scene information even when queries concern only a single instance, making them inefficient for interactive systems.

In real-world applications where users demand iterative interaction with a specific set of objects of interest, each query must be processed efficiently to provide immediate feedback. Users often need to identify or interact with objects right after reconstruction, rather than waiting through hours of preprocessing or optimization. To characterize this latency, we introduce the concept of \textbf{Time-to-First-Query (TTFQ)} -- the elapsed time between loading a reconstructed vanilla 3D scene and enabling the first open-vocabulary query and segmentation results. 
A low TTFQ is critical for responsive, scalable, and user-centric 3D systems, where fast turnaround directly translates to better usability and deployment feasibility.

Motivated by the need to minimize TTFQ for instant responsiveness, we propose a \textbf{training-free query-wise framework} for open-vocabulary 3D segmentation on Gaussian Splatting scenes. Designed with high flexibility architecture, the framework can seamlessly integrate different vision-language and video segmentation backbones without retraining, making it adaptable to evolving vision foundation models without structural modification. Given a reconstructed 3D scene represented by Gaussian primitives, we perform 3D instance segmentation from a text query without requiring any additional training. Particularly, our method directly localizes and propagates relevant 2D regions associated to the text query across camera views of the scene. These propagated masks are then lifted into 3D through a lightweight visibility voting mechanism, producing precise and complete 3D segmentation with clean boundaries under one second per query. This lightweight, training-free design circumvents the heavy preprocessing required by previous pipelines and enables interactive open-vocabulary 3D segmentation on resource-limited use cases. 
More importantly, we demonstrate that such query processing leads to state-of-the-art segmentation accuracy while preserving low query latency. 


Our contributions are:
\begin{itemize}
    \item A novel query-wise approach for open-vocabulary 3D instance segmentation in Gaussian Splatting scenes, shifting from heavy precomputing scene-level semantics to lightweight precomputing and focusing on on-demand query-driven segmentation.

   \item A dense pixel-level CLIP features that preserve fine-grained semantics beyond mask-level embeddings, enabling flexible localization of arbitrary object regions without predefined masks or feature distillation.

    \item A novel and robust multi-view 2D instance localization strategy that maps text queries to image regions via a convolutional flood-fill algorithm and propagates them across views for geometrically consistent, spatially complete 3D segmentation.

    \item A lightweight visibility voting mechanism that aggregates multi-view 2D instance regions to construct an occlusion-aware 3D instance mask, enabling real-time rendering while maintaining high-quality segmentation with minimal computational overhead.
\end{itemize}
  
\section{Related Work}
\label{sec:relatedworks}
\myheading{Visual Foundation Models (VFMs)} have greatly advanced open‑vocabulary recognition and dense perception. CLIP~\cite{clip} provides strong open-vocabulary understanding, while DINO~\cite{dino} improves dense visual representations and Grounding DINO~\cite{groundingdino} combines them for phrase-level grounding.
Other works~\cite{owlvit,fu2025llmdet} extend VLMs to detection and reasoning. However, many such methods alter CLIP during pretraining, reducing compatibility with existing open‑vocabulary systems. ProxyCLIP~\cite{proxyclip} and DeCLIP~\cite{wang2025declip} instead improve dense prediction while preserving CLIP compatibility.
For video object segmentation, SAM~\cite{sam}, SAM 2~\cite{sam2}, and HQ‑SAM~\cite{samhq} provide strong baselines for robust segmentation and tracking. Efficiency‑focused variants~\cite{edgetam, efficienttam, fastsam} further accelerate SAM‑based tracking with lightweight designs.
Leveraging advances in vision–language models and video segmentation, we build a framework that integrates efficient open‑vocabulary models with tracking‑based segmentors and 3D cues to achieve precise and temporally consistent 2D instance localization.

\myheading{Open-Vocabulary 3D Instance Segmentation} was first explored using dense point clouds~\cite{takmaz2023openmask3d,nguyen2023open3dis, nguyen_any3dis_2024, openscene, he_segpoint_2024, conceptgraph}. These approaches typically operate on RGB-D data and have demonstrated strong open-vocabulary understanding capabilities. However, the growing adoption of radiance-field-based scene representations has motivated the development of methods that leverage multi-view RGB images to achieve richer visual appearance and more efficient scene reconstruction. Neural Radiance Fields (NeRF)~\cite{nerf,mipnerf,mipnerf360,instantngp} and Gaussian Splatting~\cite{3dgs,2dgs, yonosplat, anysplat, fastgs} enable high‑quality 3D reconstruction from multi‑view RGB images. However, their radiance-field-based representations differ substantially from point clouds, introducing a distribution gap that motivates dedicated methods for open-vocabulary 3D understanding in these reconstruction frameworks. Early NeRF- and Gaussian-based approaches~\cite{lerf,langsplat,legaussian,opengaussian,instancegaussian,kim_garfield_2024,bhalgat_n2f2_2024,opensplat3d,langsplatv2} formulate segmentation as a scene-level optimization problem, performing instance and language feature learning by aggregating information from thousands of 2D masks into 3D primitives, requiring hours of preprocessing and long training cycles. Efforts to reduce training cost include Dr. Splat~\cite{drsplat}, which proposes training-free feature registration but still incurs substantial preprocessing overhead. 
VALA~\cite{vala} introduces visibility-aware feature aggregation to accelerate optimization but still requires costly scene-level semantic preprocessing.
LUDVIG~\cite{ludvig} extracts patch‑wise CLIP features and lifts them via global optimization, resulting in time‑to‑first‑query above 10 minutes and suboptimal masks due to the lack of clean supervision. B3Seg~\cite{b3seg} uses active view selection and iterative Beta–Bernoulli updates, but its reliance on instance position for camera sampling limits open-vocabulary flexibility, while its ~12-second query latency hinders real-time use.

Open-vocabulary 3D instance segmentation methods on neural primitives can be broadly categorized into pixel-based and point-based approaches. Pixel-based methods \cite{lerf,langsplat, kim_garfield_2024, bhalgat_n2f2_2024, langsplatv2, legaussian} operate in image space, where features become meaningful only after rasterization to a target view. In contrast, point-based methods \cite{opengaussian, instancegaussian, drsplat, objectgs, gaussiangrouping, segmentthensplat, ludvig} represent instance or language features directly in 3D space, enabling explicit instances and interactive manipulation.
The query mechanisms of the two paradigms also differ. Pixel-based methods must re-query instances for each viewpoint, whereas point-based methods query instances directly in 3D and render them efficiently via Gaussian rasterization. In scenarios where the primary objective is interactive 3D manipulation and fast query or inference, particularly as the number of frames increases, pixel-based methods become less efficient than point-based approaches.


Another research direction focuses on object‑aware reconstruction, segmenting instances before optimizing Gaussian scenes~\cite{gaussiangrouping, objectgs, segmentthensplat}. These methods group point clouds using 2D instance segments and avoid mask-wise language feature extraction, enabling minutes of preprocessing. Yet, they require over two hours of training per scene and lack true open‑vocabulary support, as their reliance on single‑view Grounded SAM~\cite{groundedsam} struggles with occlusions and limits general 3D applicability.
Recognizing these limitations, our method follows the point-based paradigm, eliminating both heavy preprocessing and additional training while preserving high segmentation quality and low query latency. Furthermore, it supports full open-vocabulary querying, making it well suited for interactive 3D applications and resource-limited platforms.

\section{Method}
\begin{figure*}[t]
    \centering
    \includegraphics[width=.8\linewidth]{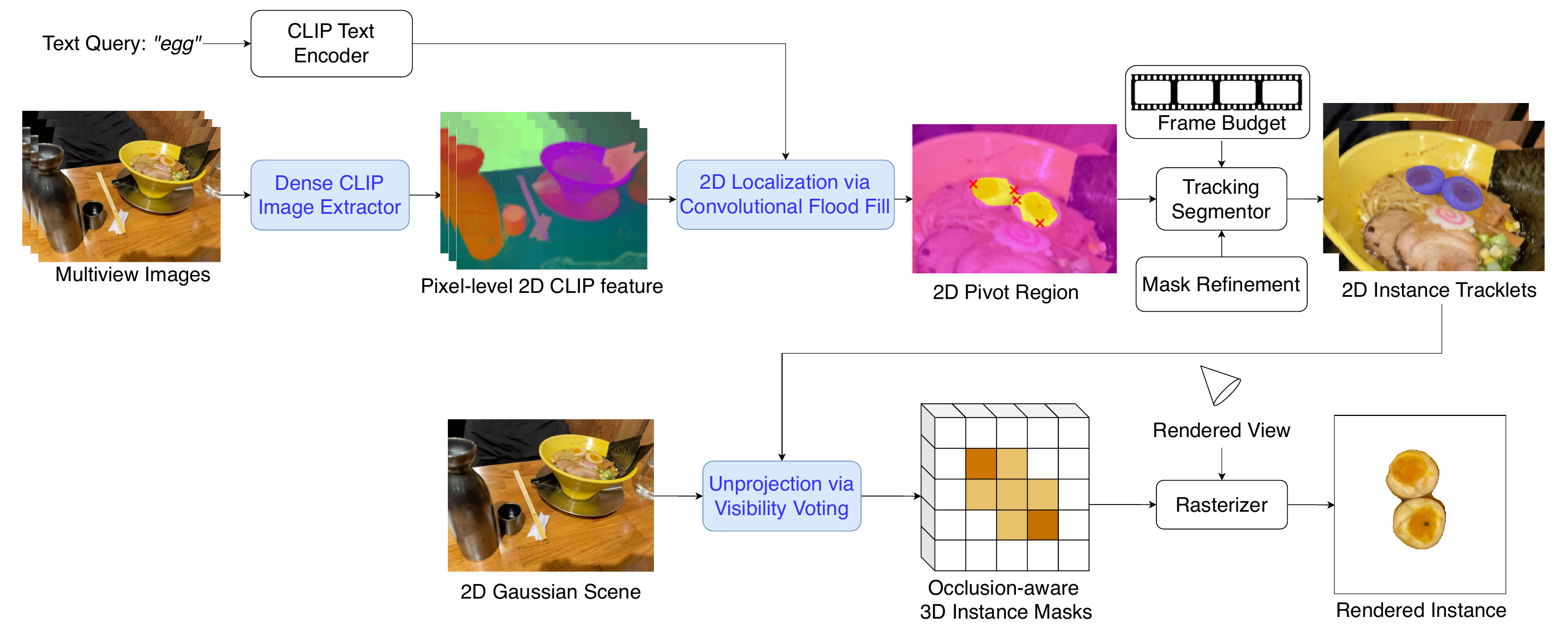}
    \caption{
        \textbf{Overview of our approach.}
        We present a novel query-based framework for open-vocabulary 3D instance segmentation. Given multi-view images and a text query, we extract pixel-level CLIP features to produce similarity maps and identify a 2D pivot instance region. The pivot is refined and propagated across views within a fixed frame budget using a Tracking Segmentor with our Mask Refinement module to ensure geometric consistency and coverage. The resulting masks are lifted to 3D Gaussians via visibility voting, forming an occlusion-aware 3D instance mask, which is rendered at novel views to obtain high-quality 2D instance segmentation.
    }
    \label{fig:2-architecture}
    \vspace{-10pt}
\end{figure*}

Our overall architecture is shown in \cref{fig:2-architecture}. Our method takes multi-view images and a text query as input, producing a 3D instance mask and rendered 2D instance masks as output. We first describe how we perform query-driven 2D instance localization directly from pixel-level CLIP features in \cref{method-localization}. Then, we explain the process of obtaining occlusion-aware 3D instance mask from these 2D regions through multi-view propagation and visibility voting in \cref{method-lifting}. Lastly, we present the real-time rendering of 2D instance masks from the constructed 3D instance mask in \cref{method-rendering}.

\subsection{Query-driven 2D Instance Localization with Dense CLIP Feature}
\label{method-localization}

Given multi‑view images and a text query, our goal is to efficiently extract the 2D pivot segment that best aligns with the query. Unlike prior methods that rely on SAM-based mask representations and are therefore constrained by predefined object proposals, we notice that dense pixel-level representations enable flexible query-driven localization of arbitrary regions, ranging from object parts and subparts to complete object instances. We further observe that, by operating directly in the image space, these representations preserve the original pixel-level vision–language alignment of CLIP features while avoiding the semantic degradation that can arise when features are distilled, compressed, or lifted into scene-level 3D representations. Moreover, our approach is inherently image-wise: feature extraction is performed only once per image, after which arbitrary queries can be evaluated directly on the dense feature maps. This is substantially more efficient than mask-wise pipelines~\cite{legaussian,opengaussian,langsplat,drsplat,instancegaussian,langsplatv2}, which require processing and scoring a large number of mask proposals for every image.

We begin by computing dense pixel‑level CLIP features for each image using ProxyCLIP~\cite{proxyclip}, a training‑free method that aligns CLIP embeddings with spatial VFM features for dense vision–language correspondence. The text query is encoded into a CLIP text embedding, after which we compute cosine similarity between the image features
$\mathbf{F}_{img} \in \mathbb{R}^{N \times H \times W \times D}$
and the query embedding
$\mathbf{F}_{query} \in \mathbb{R}^{D}$
to obtain similarity maps
$\mathbf{S} \in \mathbb{R}^{N \times H \times W}$,
where $N$, $H$, $W$, and $D$ denote the number of views, image height, width, and feature dimension, respectively.

\begin{figure}[t]
    \centering
    \includegraphics[width=0.7\columnwidth]{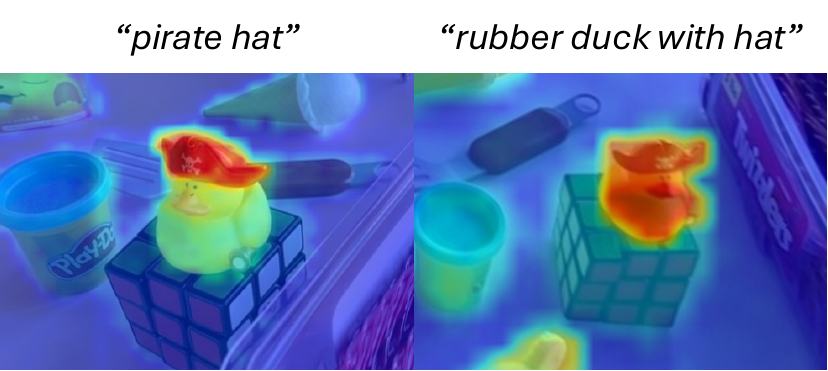}
    \caption{Cosine similarity map of the same object under two distinct queries.}
    \label{fig:3-cosine_similarity}
    \vspace{-10pt}
\end{figure}

After computing pixel-level cosine similarity maps, we localize the instance most relevant to the text query by selecting the highest‑scoring pixel within the central image region across all views, ensuring that the chosen location corresponds to a clearly visible instance. A naive approach would use this pixel as a point prompt for an instance segmentor, but single‑pixel prompting often yields ambiguous masks--capturing only part of the object or leaking into surrounding regions. As shown in \cref{fig:3-cosine_similarity}, cosine similarity maps contain rich high‑response areas that better indicate candidate pivot regions. To extract these regions, we threshold the similarity map with $\tau$ to obtain a binary mask, then remove small noisy components using a minimum-area filter $A_{\min}$. Starting from the top‑scoring pixel, we apply a convolutional flood‑fill algorithm~\cite{floodfill} to grow the connected region, followed by convolutional morphological erosion~\cite{morphologicalerosion} to refine it into a clean, compact pivot mask.

\subsection{Occlusion-aware 3D Instance Mask Formation via Visibility Voting}
\label{method-lifting}
The localized 2D pivot region from Sec.~\ref{method-localization} serves as the initial instance hypothesis. We propagate it across visible frames using a tracking segmentor to recover the complete object, enabling geometrically consistent and spatially complete 3D segmentation.

\myheading{Budget-aware frame selection and ordering.}
Selecting a strong pivot view is crucial for high-quality 3D segmentation. The pivot region, identified as the most query-relevant region in the most visually prominent view, provides the richest semantic and spatial cues for the target object. 
%
Propagating this region to all views is unnecessary and inefficient, we propose a lightweight view-selection protocol that filters out irrelevant frames while preserving complete coverage of the target instance. We denote the 3D Gaussians whose projections lie within the pivot region as pivot Gaussians \(\mathcal{G}_{\text{pivot}}\), as they encode reliable geometric and semantic cues from the object’s most representative viewpoint. For each pivot region obtained in the pivot frame $\mathcal{I}_{\text{pivot}}$ with index $i$ (referred to as frame $i$), we identify candidate frames in which the pivot Gaussians \(\mathcal{G}_{\text{pivot}}\) are visible. Let $n_j$ denotes the ratio of projection of pivot Gaussians \(\mathcal{G}_{\text{pivot}}\) visible in frame $j$. Given a budget of $P$ frames (excluding the pivot frame), where $P$ is carefully chosen to balance instance coverage and propagation latency, we sample frames along a 360-degree camera trajectory around the target instance and select up to $P/2$ temporally preceding and $P/2$ temporally succeeding frames relative to the pivot frame according to the largest $n_j$ values, or:
\begin{equation}
\mathcal{I}_{\text{back}} = \text{Top}\big(\{n_j\}_{j < i}, P/2\big), \quad
\mathcal{I}_{\text{fwd}} = \text{Top}\big(\{n_j\}_{j > i}, P/2\big).
\label{eq:frame-select}
\end{equation}
Backward frames $\mathcal{I}_{\text{back}}$ are arranged in reverse temporal order, whereas forward frames $\mathcal{I}_{\text{fwd}}$ follow ascending order. The pivot frame is prepended to both sequences, which are then concatenated (denoted by $[\ldots; \ldots; \ldots]$) to form a temporally coherent image stream, thereby improving the stability and performance of the tracking process:
\begin{equation}
\mathcal{I} = [\mathcal{I}_{\text{pivot}}; \mathcal{I}_{\text{back}}; \mathcal{I}_{\text{pivot}}; \mathcal{I}_{\text{fwd}}].
\label{eq:frame-order}
\end{equation}

After determining the frames for propagation, we leverage a tracking segmentor \cite{sam2, samhq, edgetam, efficienttam} to propagate the pivot region from the reference frame to all candidate frames, obtaining the instance tracklet $\mathbf{M} = {\mathbf{m}_1, \mathbf{m}_2, \ldots, \mathbf{m}_{P+1}}$.

\myheading{Mask Refinement.}
Although this ordering enforces bidirectional temporal expansion from the pivot frame, promoting continuity in views where the instance remains visible, propagation errors may sometimes occur due to the performance limitations of the tracking segmentor, particularly under significant viewpoint changes between adjacent frames or when the instance is confused with visually similar objects. To mitigate propagation errors, we utilize pivot Gaussians \(\mathcal{G}_{\text{pivot}}\) as reliable 3D geometric cues to regularize and stabilize the tracking process. For each tracklet, let \(\mathcal{G}_{\text{out}}\) denote the subset of pivot Gaussians whose projections fall outside the tracklet. If
$
\frac{|\mathcal{G}_{\text{out}}|}{|\mathcal{G}_{\text{pivot}}|} > 0.5,
$
the tracklet is considered misaligned. Following the prompting mechanism of SAM-based tracking segmentors~\cite{sam2}, we apply Farthest Point Sampling (FPS) on tracklet pixels to select negative prompts and on the projected visible pivot Gaussians $\mathcal{G}_{\text{pivot}}$ in the current view as positive prompts to refine the mask.

\myheading{Lifting to 3D via Visibility Voting.}
A straightforward approach to construct a 3D instance mask is projection mapping, assigning a Gaussian to the instance if its projection falls within the tracked 2D masks. However, the unstructured nature of Gaussian primitives and inevitable tracking errors make this approach highly sensitive to incorrect tracklets. To address this sensitivity, we introduce a visibility‑aware voting mechanism that aggregates evidence across views to robustly decide each Gaussian’s instance membership. The key idea is to treat erroneous segments as outliers: even if a Gaussian is misassigned in one view, consistent assignments in other views can correct it through multi‑view voting.

%
For a video sequence $\mathcal{I}$, we project 3D Gaussians into each view $i \in \mathcal{I}$ and evaluate them against the propagated 2D instance tracklets $\mathbf{M}$ to enforce cross-view consistency. At each frame $i$, we derive the mask $\mathbf{m}_i \in \mathbf{M}$ and the Gaussian visibility indicator $\mathbf{V}_i \in \{0, 1\}^N$, where $N$ is the total number of Gaussians. The visibility indicator specifies whether each Gaussian is visible or occluded in the current frame. In each view where it is visible, a Gaussian casts a foreground vote if its projection lies within the mask $\mathbf{m}_i$, and a background vote otherwise. The per-Gaussian foreground and background frequencies (both $\in \mathbb{N}^N$) are then aggregated across all views as follows:
\begin{equation}
\begin{aligned}
f_g^{\text{fg}}
&=
\sum_{i \in \mathcal{I}}
\mathbf{m}_i[\mathbf{P}_i(g)]
\cdot
\mathbf{V}_i(g), \\
f_g^{\text{bg}}
&=
\sum_{i \in \mathcal{I}}
\left(1-\mathbf{m}_i[\mathbf{P}_i(g)]\right)
\cdot
\mathbf{V}_i(g).
\end{aligned}
\label{eq:visibility-voting-1}
\end{equation}
Here, instead of storing the binary 3D instance mask, we augment Gaussian Splatting with a compact, integer-valued occlusion-aware 3D instance mask, facilitating high-quality instance rendering:
\begin{equation}
\mathcal{F}_{\text{inst}} = 
\begin{bmatrix}
f^{\text{fg}};f^{\text{bg}}
\end{bmatrix} 
\in \mathbb{Z}^{N \times 2}.
\label{eq:instance-feature}
\end{equation}

The constructed 3D instance mask offers a comprehensive representation that enables accurate, occlusion-aware instance segmentation in every view. Memory efficiency is achieved because the foreground-background indicators are mostly zeros and can be stored compactly using a sparse, integer-based format.

\myheading{Multi-instance Segmentation.}  
For queries corresponding to multiple instances, we segment them iteratively. In each iteration, the top-scoring pixel in the 2D pivot region defines the instance score. After extracting an instance, we propagate its Gaussians $\mathcal{G}_{\text{inst}}$ to all visible views and suppress the cosine similarity of pixels influenced by these Gaussians to the minimum value. This removes the segmented instance from subsequent iterations, enabling efficient extraction of multiple instances for the same query.

\subsection{Real-Time Gaussian Semantic Rasterizer}
\label{method-rendering}
We adopt a rasterization process similar to Gaussian Splatting \cite{2dgs,3dgs} to render instance masks in real time. Specifically, for color rendering, a screen-space bounding box is first computed for each Gaussian primitive. The Gaussians are then depth-sorted and organized into tiles based on their bounding boxes. Finally, volumetric alpha blending is applied to integrate the alpha-weighted appearance from front to back:

\begin{equation}
\begin{aligned}
\mathcal{F'}_{\text{inst}}(\mathbf{x})
&=
\sum_{i=1}^{N}
\mathcal{F}_{\text{inst}}[i]
\, \alpha_i
\prod_{j=1}^{i-1}
\bigl(1-\alpha_j\bigr), \\
\mathcal{F'}_{\text{inst}}
&=
[f'^{\text{fg}},\, f'^{\text{bg}}]
\in \mathbb{R}^{H \times W \times 2}.
\end{aligned}
\label{eq:gaussian-feature-render}
\end{equation}

During rendering, $f^{\text{fg}}$ and $f^{\text{bg}}$ are composited following the standard color-opacity blending process, yielding the per-pixel instance visibility map $\mathcal{F'}_{\text{inst}}$. Pixels where $f'^{\text{fg}} > f'^{\text{bg}}$ define the instance masks, a lightweight operation that substantially enhances visual quality. In the multi-instance setting, each instance is associated with a cosine similarity score reflecting its relevance to the query. At each view, the visible instance with the highest similarity is prioritized, ensuring accurate handling of overlapping regions while preserving rasterization efficiency.

\begingroup
\renewcommand{\thefootnote}{*}
\section{Experiments\protect\footnotemark}
\footnotetext{Please refer to the \textcolor{blue}{Supplementary Material} for additional 3D/2D qualitative results and implementation details.}
\endgroup

\begin{table}[t]
\centering
\caption{Open-vocabulary 3DIS on LERF-OVS~\cite{lerf}.}
\label{tab:lerf-ovs}
\vspace{-5pt}
\resizebox{.95\columnwidth}{!}{%
\begin{tabular}{lccc}
\toprule
Method & mIoU $\uparrow$ & mAcc@0.25 $\uparrow$ & TTFQ $\downarrow$ \\
\midrule
\multicolumn{4}{l}{\textit{Training-based \& Pixel-level}} \\
LERF \cite{lerf} & 37.48 & 76.88 & 34 min \\
LangSplat \cite{langsplat} & 46.75 & 68.13 & 7 hours \\
LEGaussians \cite{legaussian} & 46.63 & 59.91 & 7 hours \\
LangSplatV2 \cite{langsplatv2} & \underline{59.90} & \underline{84.10} & 10 hours \\
\midrule
\multicolumn{4}{l}{\textit{Training-based \& Point-level}} \\
OpenGaussian \cite{opengaussian} & 38.36 & 51.43 & 7 hours \\
InstanceGaussian \cite{instancegaussian} & 43.63 & 62.82 & 7 hours \\
Gaussian Grouping \cite{gaussiangrouping} & 43.16 & 50.65 & 1 hour \\
ObjectGS \cite{objectgs} & 50.08 & 60.64 & 2 hours \\
Segment-then-Splat \cite{segmentthensplat} & 52.10 & -- & 2 hours \\
VALA~\cite{vala} & 58.02 & 82.85 & 6 hours \\
\midrule
\multicolumn{4}{l}{\textit{Training-free \& Point-level}} \\
Dr. Splat \cite{drsplat} & 43.58 & 63.87 & 6 hours \\
LUDVIG \cite{ludvig} & 50.58 & 60.59 & \underline{10 min} \\
Ours & \textbf{63.02} & \textbf{84.91} & \textbf{1.5 min} \\
\bottomrule
\end{tabular}
}
\vspace{-10pt}
\end{table}

\begin{table}[t]
\centering
\scriptsize
\caption{Open-vocabulary 3DIS on 3D-OVS \cite{3dovs}.}
\resizebox{\columnwidth}{!}{
\label{tab:3d-ovs}
\begin{tabular}{lcccccc}
\toprule
\multirow{2}{*}{Method} & \multicolumn{5}{c}{mIoU $\uparrow$} & \multirow{2}{*}{Mean} \\
\cmidrule(lr){2-6}
 & bed & bench & lawn & room & sofa &  \\
\midrule
\multicolumn{7}{l}{\textit{Training-based \& Pixel-level}} \\
LERF \cite{lerf} & 73.50 & 53.20 & 74.00 & 46.60 & 73.70 & 64.20 \\
LangSplat \cite{langsplat} & 77.80 & 77.30 & 90.90 & 58.40 & 60.20 & 72.92 \\
LEGaussian \cite{legaussian} & 44.57 & 59.91 & 57.31 & 39.96 & 40.89 & 48.53 \\
LangSplatV2 \cite{langsplatv2} & 93.00 & \underline{94.90} & 96.60 & \textbf{96.10} & \underline{92.30} & 94.60 \\
\midrule
\multicolumn{7}{l}{\textit{Training-based \& Point-level}} \\
3D-OVS \cite{3dovs} & 89.50 & 89.30 & 88.20 & 92.80 & 74.00 & 86.76 \\
OpenGaussian \cite{opengaussian} & 28.42 & 49.45 & 41.20 & 24.63 & 41.20 & 36.98 \\
InstanceGaussian \cite{instancegaussian} & 55.05 & 71.59 & 52.46 & 78.71 & 65.85 & 64.73 \\
Gaussian Grouping \cite{gaussiangrouping} & 64.50 & 95.60 & 97.00 & 96.40 & 91.30 & 88.96 \\
Segment-then-Splat \cite{segmentthensplat} & -- & -- & -- & -- & -- & 88.53 \\
ObjectGS \cite{objectgs} & \textbf{98.00} & \textbf{96.40} & \textbf{97.20} & 95.10 & \textbf{95.40} & \textbf{96.42} \\
\midrule
\multicolumn{7}{l}{\textit{Training-free \& Point-level}} \\
Ours & \underline{95.25 }& 94.45 & \underline{96.76} & \underline{95.53} & 91.47 & \underline{94.69} \\
\bottomrule
\end{tabular}
} 
\vspace{-10pt}
\end{table}

\begin{table}[h]
\centering
\scriptsize
\caption{Open-vocabulary 3DIS on ScanNet \cite{scannet}.
}
\label{tab:scannet}
\begin{tabular}{lccc}
\toprule
Method 
& mIoU $\uparrow$ 
& mAcc@0.25 $\uparrow$
& TTFQ $\downarrow$ \\
\midrule
LUDVIG \cite{ludvig}       & 23.89 & 51.85 & 12 mins \\
LangSplatV2 \cite{langsplatv2}   & 39.44 & 53.06 & 10 hours \\
\midrule
Ours          & \textbf{43.74} & \textbf{53.61} & \textbf{2 mins} \\
\bottomrule
\end{tabular}
\vspace{-10pt}
\end{table}

\begin{figure}[t]
    \centering
    \includegraphics[width=\columnwidth]{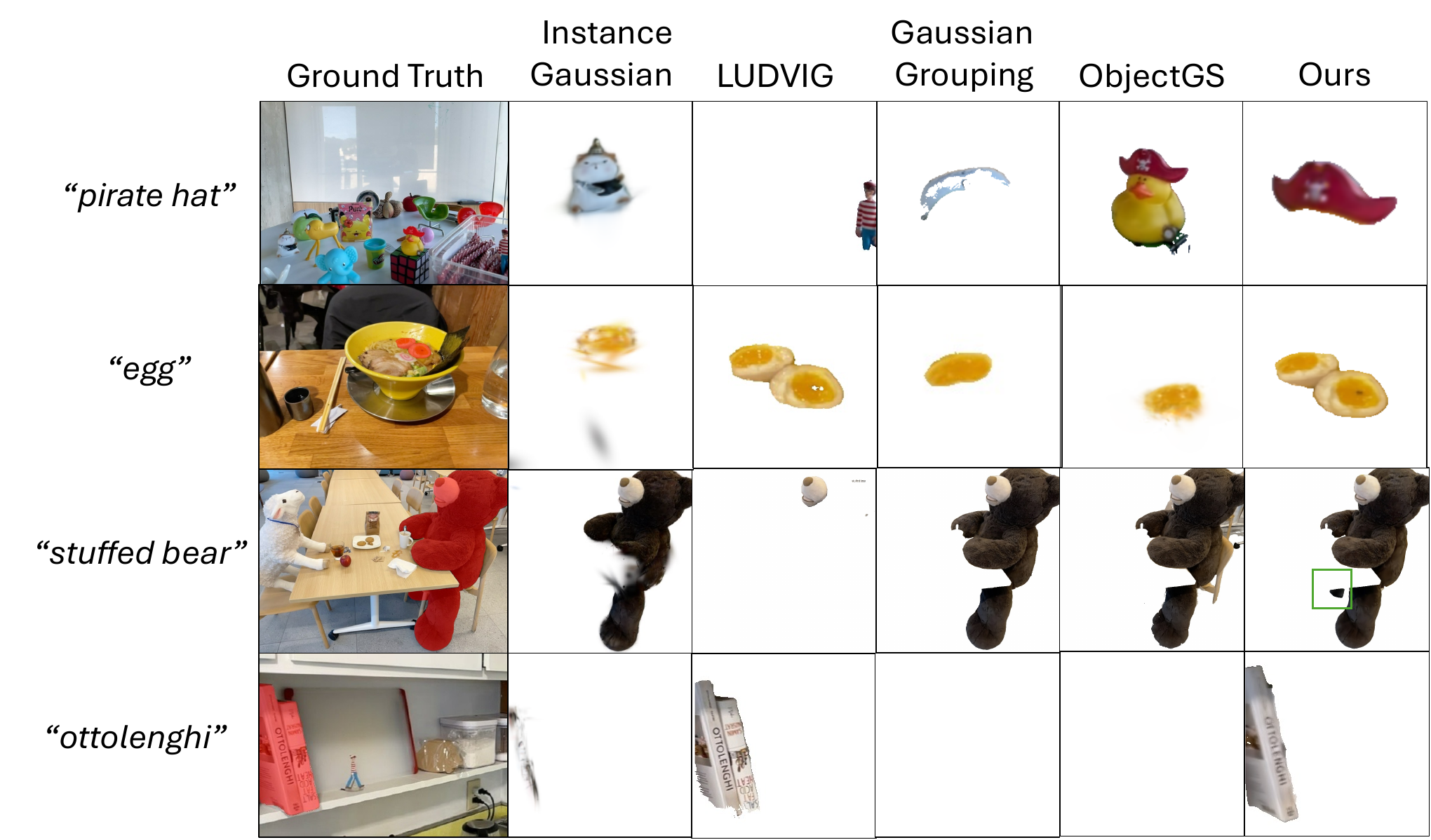}
    \vspace{-5pt}
    \caption{
    Qualitative results on LERF-OVS \cite{lerf}. 
    }
    \label{fig:6-lerf_ovs}
    \vspace{-10pt}
\end{figure}

\begin{figure*}[t]
    \centering
    \includegraphics[width=0.8\linewidth]{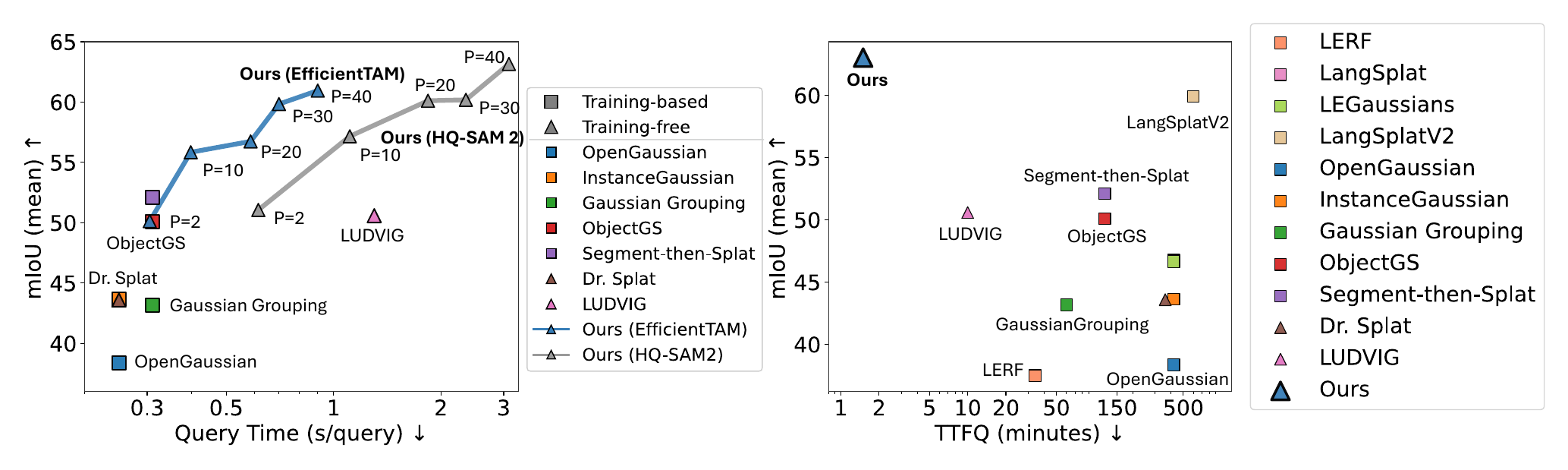}
    \vspace{-5pt}
    \caption{Accuracy–efficiency trade-off comparison on LERF-OVS \cite{lerf}.}
    \label{fig:9-query-ttfq}
    \vspace{-10pt}
\end{figure*}

\begin{figure}[t]
    \centering
    \includegraphics[width=\columnwidth]{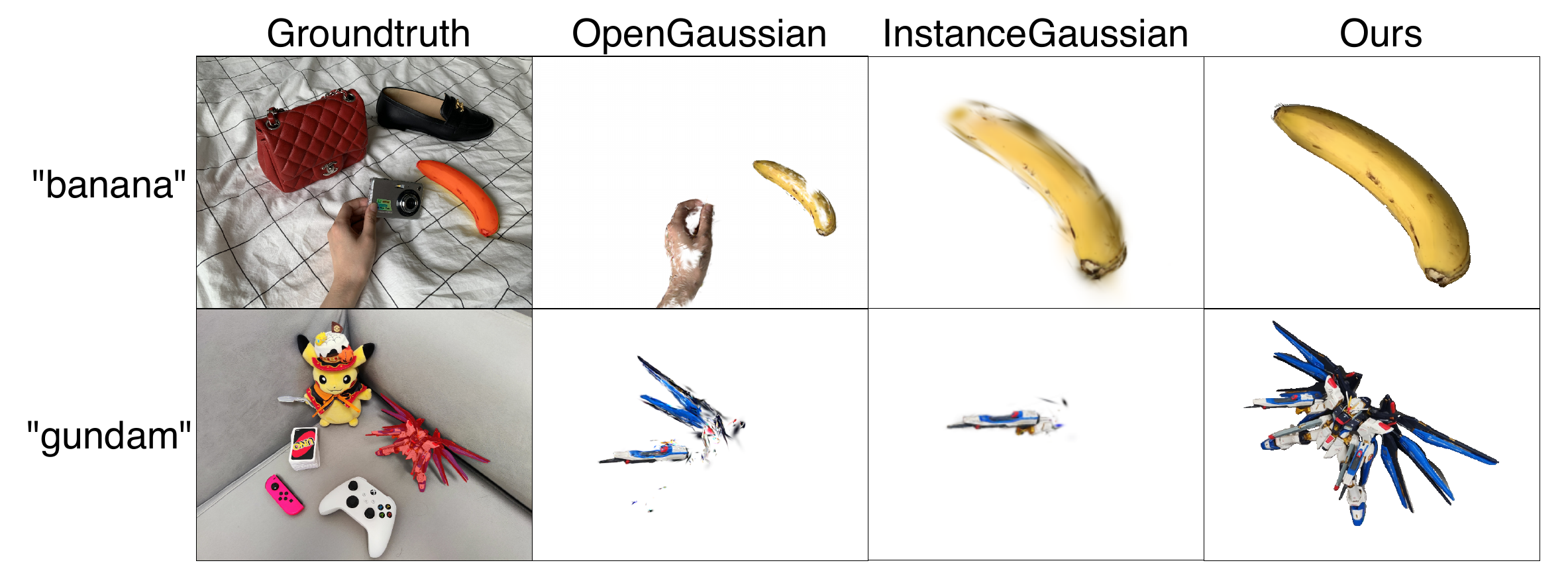}
    \vspace{-5pt}
    \caption{
    Qualitative results on 3D-OVS \cite{3dovs}
    }
    \label{fig:7-3d-ovs}
    \vspace{-10pt}
\end{figure}

\subsection{Open-Vocabulary 3D Instance Segmentation}
\myheading{Setting:} Following prior work~\cite{legaussian,lerf,langsplat,langsplatv2,opengaussian,ludvig,drsplat}, we evaluate open-vocabulary 3D instance segmentation using the standard rendered-2D-versus-ground-truth-2D protocol, where 3D predictions are rendered into target views and compared against annotated 2D instance masks. Given a text query, each method encodes it with the CLIP text encoder, matches it to 3D instances via cosine similarity to retrieve the top-scoring instance, and renders it across predefined viewpoints using Gaussian rasterization. Time‑to‑First‑Query (TTFQ) measures the time from loading a reconstructed vanilla 3D Gaussian/NeRF scene to generating the first query result. Query latency captures the per‑query response time. For fair comparison, we report query latency only among point‑based methods, as our approach is also point‑based and operates directly at the 3D level.
\myheading{Datasets:} 
We evaluate our method on three standard open-vocabulary 3D instance segmentation benchmarks: LERF-OVS~\cite{lerf}, 3D-OVS~\cite{3dovs}, and ScanNet~\cite{scannet}. To assess performance in large-scale, densely cluttered environments, we further introduce new 2D segmentation annotations for ScanNet, covering 63 categories, enabling evaluation of unique-instance localization under challenging scene complexity.
\myheading{Result:} 
\textbf{Quantitative results} are presented in \cref{tab:lerf-ovs} for LERF‑OVS, \cref{tab:3d-ovs} for 3D‑OVS, and \cref{tab:scannet} for ScanNet, with \textbf{qualitative visualizations} shown in \cref{fig:6-lerf_ovs}, \cref{fig:9-query-ttfq}, and \cref{fig:7-3d-ovs}. Across both conventional (LERF-OVS) and cluttered (ScanNet) scenes, our method achieves state-of-the-art accuracy while substantially reducing TTFQ and maintaining low query latency, outperforming lifting-based, optimization-heavy, and instance-aware reconstruction methods such as ObjectGS~\cite{objectgs}, Gaussian Grouping~\cite{gaussiangrouping}, and LangSplatV2~\cite{langsplatv2}, all while operating with far lower preprocessing cost. The slightly lower 3D-OVS scores stem from vanilla 3DGS's unstructured spatial layout, as our method uses no instance-aware pretraining. In contrast, ObjectGS requires $\sim$2 hours of instance-aware 3DGS training per scene, while ours is training-free with only 1.5 minutes of preprocessing. On ObjectGS's instance-aware 3DGS, our method achieves \textbf{96.55} mIoU (vs. 94.69 vanilla, 96.42 ObjectGS), confirming that the gap primarily stems from the representation.

\subsection{Open-Vocabulary Point Cloud Understanding}
\myheading{Setting:} 
Although designed for query-wise 3D instance segmentation, our framework naturally extends to full-scene understanding with lightweight modifications while preserving its core design and training-free nature. We first generate class-agnostic 3D proposals, then assign the average tracklet feature to each proposal's Gaussians. We evaluate class-agnostic 3D instance segmentation on the full validation split of ScanNet++~\cite{scannetpp} following the exact Open3DIS protocol without any post-processing. We further evaluate semantic understanding on ScanNet~\cite{scannet} using the OpenGaussian evaluation protocol. Implementation details are provided in the supplementary material.
\myheading{Datasets:} We evaluate class-agnostic instance segmentation on the full ScanNet++~\cite{scannetpp} validation split following Open3DIS~\cite{nguyen2023open3dis} and semantic segmentation on 10 ScanNet~\cite{scannet} scenes following OpenGaussian~\cite{opengaussian}.
\myheading{Results:}
Our adapted pipeline generalizes well beyond the query-wise setting. On ScanNet++, it achieves the highest AP (22.2) while remaining competitive on other metrics (\cref{tab:comparison_scannetpp}). On ScanNet, it establishes new state of the art, outperforming prior methods across all class splits in both mIoU and mAcc (\cref{tab:semantic_seg}). These gains confirm the versatility and robustness of our core framework, showing that it not only excels in efficient query-wise open-vocabulary segmentation but also delivers competitive performance in full-scene class-agnostic and semantic tasks with minimal adaptation.
\begin{table}[t]
\small
\centering
\caption{Results of class-agnostic 3DIS on ScanNet++~\cite{scannetpp}}
\label{tab:comparison_scannetpp}
\resizebox{0.6\linewidth}{!}{
\begin{tabular}{l ccc}
\toprule
Method & AP$\uparrow$ & AP$_{50}$$\uparrow$ & AP$_{25}$$\uparrow$ \\
\midrule
Segment3D~\cite{huang_segment3d_2023} & 12.0 & 22.7 & 37.8 \\
Open3DIS~\cite{nguyen2023open3dis} & 20.7 & \textbf{38.6} & 47.1 \\
OpenSplat3D~\cite{opensplat3d} & 19.2 & 37.3 & \textbf{56.2} \\
\midrule
\textbf{Ours} & \textbf{22.2} & 35.8 & 47.2 \\
\bottomrule
\end{tabular}}
\end{table}
\begin{table}[t]
\small
\centering
\caption{Results of semantic segmentation on ScanNet.}
\label{tab:semantic_seg}
\resizebox{0.9\linewidth}{!}{
\begin{tabular}{l|cc|cc|cc}
\toprule
\multirow{2}{*}{Methods} &
\multicolumn{2}{c|}{19 classes} &
\multicolumn{2}{c|}{15 classes} &
\multicolumn{2}{c}{10 classes} \\
& mIoU$\uparrow$ & mAcc$\uparrow$ & mIoU$\uparrow$ & mAcc$\uparrow$ & mIoU$\uparrow$ & mAcc$\uparrow$ \\
\midrule
LangSplat~\cite{langsplat} & 3.8 & 9.1 & 5.4 & 13.2 & 8.4 & 22.1 \\
LEGaussians~\cite{legaussian} & 3.8 & 10.9 & 9.0 & 22.2 & 12.8 & 28.6 \\
OpenGaussian~\cite{opengaussian} & 24.7 & 41.5 & 30.1 & 48.3 & 38.3 & 55.2 \\
VALA~\cite{vala} & 32.1 & 50.0 & 35.1 & 54.7 & 46.2 & 65.6 \\
LUDVIG~\cite{ludvig} & 33.9 & 51.4 & 37.4 & 57.2 & 46.4 & 66.2 \\
\midrule
\textbf{Ours} & \textbf{38.9} & \textbf{62.3} & \textbf{40.0} & \textbf{63.1} & \textbf{50.3} & \textbf{71.5} \\
\bottomrule
\end{tabular}
}
\end{table}
%

\subsection{Computing Device Analysis}
We evaluate our method using EfficientTAM~\cite{efficienttam} with $P=40$ on three different GPUs. Our method is entirely \emph{training-free} and can be efficiently deployed on consumer-grade hardware. As shown in Table 3 of our Supp. and ~\Cref{tab:computing-device}, our approach is the \textbf{only} method among compared baselines that successfully runs on an RTX 3060 8GB laptop GPU without encountering out-of-memory (OOM) errors. This efficiency stems from its low memory footprint, requiring only $\sim$0.68GB for feature storage and a peak VRAM of 5GB throughout the entire pipeline. In contrast, other methods~\cite{langsplatv2,objectgs,ludvig,opengaussian,instancegaussian,lerf,legaussian,drsplat} fail to run under the same memory constraints due to the substantial overhead of feature distillation and scene-specific optimization. Despite operating on resource-limited hardware, our method achieves a Time-to-First-Query (TTFQ) of only 1.2-1.8 minutes, demonstrating its practicality for efficient open-vocabulary 3D understanding.
\begin{table}[t]
\centering
\caption{Computing device analysis on LERF-OVS using EfficientTAM ($P=40$).}
\label{tab:computing-device}
\scriptsize
\begin{tabular}{lccc}
\toprule
GPU Type & mIoU $\uparrow$ & TTFQ (min) $\downarrow$ & Query (s) $\downarrow$ \\
\midrule
Server (A100 40GB)      & 60.95 & 1.5  & 0.90 \\
Desktop (RTX 4090 24GB) & 60.72 & 1.2  & 0.47 \\
Laptop (RTX 3060 8GB)   & 60.75 & 1.8  & 1.41 \\
\bottomrule
\end{tabular}
\end{table}

\subsection{Ablation Study}
In this section, we present an ablation study of our proposed components on the LERF-OVS framework.

\myheading{Effectiveness of each proposed component.}
Since no prior baseline exists, we construct a three-stage query-wise baseline comprising (1) feature extraction, (2) multi-view localization, and (3) 3D lifting. The baseline version uses (1) mask-wise CLIP feature extraction, (2) pivot localization by cosine similarity matching of text-query CLIP with SAM mask embeddings, followed by HQ-SAM 2 multi-view propagation~\cite{samhq}, and (3) projection mapping for 3D lifting.
We then replace each stage with our proposed (1) \textbf{Dense CLIP Features}, (2) \textbf{Mask Refinement} and \textbf{Ranked Frame Selection}, and (3) \textbf{Visibility Voting}. As shown in \cref{tab:major-ablation}, each component consistently improves performance, with the full model achieving \textbf{63.02} mIoU compared to only \textbf{14.95} mIoU for the baseline pipeline despite using the same underlying ingredients. In particular, replacing projection mapping with our visibility voting mechanism yields a large improvement from \textbf{33.20} to \textbf{63.02} mIoU, highlighting the importance of occlusion-aware 3D aggregation. The substantial performance gap between the mask-based baseline and Dense CLIP Features (\textbf{35.41} vs. \textbf{63.02} mIoU), together with a significant reduction in preprocessing time (\textbf{345} vs. \textbf{1.5} minutes). This agrees with LUDVIG~\cite{ludvig}, where replacing patch-wise CLIP features with dense image-wise features improved performance (\textbf{+1.6} localization accuracy, \textbf{+17.7} mIoU without diffusion, \textbf{+0.4} mIoU with diffusion), highlighting the importance of preserving pixel-level vision-language alignment. These results suggest that the gains stem not only from stronger individual components but also from the proposed query-wise formulation itself, which fundamentally differs from prior full-scene feature learning pipelines.

\begin{table}[t]
\centering
\scriptsize
\setlength{\tabcolsep}{3pt}
\caption{Effectiveness of each component on LERF-OVS~\cite{lerf}.}
\label{tab:major-ablation}
\begin{tabular}{ccccccc}
\toprule
\multicolumn{4}{c}{Component} & \multicolumn{3}{c}{Metrics} \\
\cmidrule(lr){1-4}\cmidrule(lr){5-7}
\makecell{Dense\\CLIP} &
\makecell{Mask\\Refinement} &
\makecell{Ranked\\Frame} &
\makecell{Visibility\\Voting} &
mIoU &
\makecell{Preprocess\\(min)} &
\makecell{Query\\(s)} \\
\midrule
\checkmark & \checkmark & \checkmark & \checkmark & \textbf{63.02} & \textbf{1.50} & \textbf{3.10} \\
\midrule
\checkmark & \checkmark & \checkmark &  & 33.20 & 1.50 & 4.16 \\
\checkmark & \checkmark &  & \checkmark & 60.38 & 1.50 & 42.15 \\
\checkmark &  & \checkmark & \checkmark & 56.24 & 1.50 & 3.10 \\
 & \checkmark & \checkmark & \checkmark & 35.41 & 345 & 3.87 \\
 &  &  &  & 14.95 & 345 & 30.15 \\
\bottomrule
\end{tabular}
\vspace{-5pt}
\end{table}

\myheading{Effectiveness of different trackers.}
We compare several tracking‑based segmentors to assess their impact on query‑wise instance propagation and overall segmentation quality. As shown in \cref{tab:ablation_tracking} and \cref{fig:9-query-ttfq}, HQ‑SAM 2~\cite{samhq} achieves the highest mIoU, demonstrating superior boundary precision and consistent instance tracking across frames. EfficientTAM~\cite{efficienttam}, while slightly less accurate, offers significantly better efficiency, reducing average query time to under 1s and maintaining competitive quality. This trade‑off highlights the flexibility of our framework, allowing users to balance accuracy and runtime while naturally benefiting from continued advances in video object segmentation (VOS) methods.

\begin{table}[t]
\centering
\scriptsize
\caption{Tracking performance with different \# frames.}
\label{tab:ablation_tracking}
\vspace{-10pt}
\begin{tabular}{lccc}
\toprule
\shortstack{Tracking Segmentor} &
\shortstack{Frame Budget} &
mIoU $\uparrow$ &
\shortstack{Query Time} $\downarrow$ \\
\midrule

\multirow{5}{*}{EfficientTAM \cite{efficienttam}}
& 2  & 50.13 & \textbf{0.31} s \\
& 10 & 55.84 & 0.40 s \\
& 20 & 56.74 & 0.59 s \\
& 30 & 59.83 & 0.70 s \\
& 40 & \textbf{60.95} & 0.90 s \\
\midrule

\multirow{5}{*}{HQ-SAM 2 \cite{samhq}}
& 2  & 51.05 & \textbf{0.62} s \\
& 10 & 57.16 & 1.11 s \\
& 20 & 60.11 & 1.84 s \\
& 30 & 60.18 & 2.35 s \\
& 40 & \textbf{63.02} & 3.10 s \\
\bottomrule
\end{tabular}
\vspace{-10pt}
\end{table}

\section{Conclusion}

We presented a practical and scalable framework for open-vocabulary 3D instance segmentation that balances accuracy and efficiency. Our query-wise, training-free approach combines dense vision-language features, robust 2D localization, and visibility voting to construct occlusion-aware 3D mask to achieve state-of-the-art accuracy with low computational overhead and substantially reduced TTFQ. Comprehensive experiments across diverse and challenging benchmarks demonstrate strong generalization and compatibility with consumer-grade hardware.
%

%
%
\bibliographystyle{splncs04}
\bibliography{main}
\end{document}